\documentclass[a4paper, 10pt, conference]{ieeeconf}      
\usepackage{FG2024}
\usepackage{cite}
\usepackage{amsmath,amssymb,amsfonts}
\usepackage{graphicx}
\usepackage{textcomp}
\usepackage{xcolor}
\usepackage{algorithm}
\usepackage{algpseudocode}
\usepackage{booktabs}
\usepackage{mathabx}
\usepackage{multirow}
\usepackage{graphics}
\usepackage{adjustbox}
\usepackage{subcaption}
\usepackage[font=small,labelfont=bf]{caption}
\usepackage{url}
\usepackage{tabularx}

\FGfinalcopy 

\IEEEoverridecommandlockouts                              
\def\FGPaperID{255} 

\title{\LARGE \bf
Explainable Face Verification via Feature-Guided Gradient Backpropagation
}

\author{\parbox{16cm}{\centering
    {\large Yuhang Lu$^*$, Zewei Xu$^*$, and Touradj Ebrahimi}\\
    {
     EPFL, Lausanne, Switzerland \\
\tt\small firstname.lastname@epfl.ch}}
    \thanks{$^\ast$Equal contribution}
    \thanks{Support from XAIface CHIST-ERA-19-XAI-011 and the Swiss National Science Foundation (SNSF) 20CH21\_195532 is acknowledged.}
}

\begin{document}

\ifFGfinal
\thispagestyle{empty}
\pagestyle{empty}
\else
\author{Anonymous FG2024 submission\\ Paper ID \FGPaperID \\}
\pagestyle{plain}
\fi
\maketitle

\begin{abstract}
Recent years have witnessed significant advancement in face recognition (FR) techniques, with their applications widely spread in people's lives and security-sensitive areas. There is a growing need for reliable interpretations of decisions of such systems. Existing studies relying on various mechanisms have investigated the usage of saliency maps as an explanation approach, but suffer from different limitations. This paper first explores the spatial relationship between face image and its deep representation via gradient backpropagation. Then a new explanation approach FGGB has been conceived, which provides precise and insightful similarity and dissimilarity saliency maps to explain the ``Accept'' and ``Reject'' decision of an FR system. Extensive visual presentation and quantitative measurement have shown that FGGB achieves superior performance in both similarity and dissimilarity maps when compared to current state-of-the-art explainable face verification approaches.

\end{abstract}

\section{Introduction}
\label{sec:intro}




Over the past decades, the accuracy of Face Recognition (FR) systems has been boosted due to the advanced technologies based on deep convolutional neural networks (DCNNs) \cite{simonyan2014very, he2016deep, hu2018squeeze, tan2019efficientnet} and large-scale face datasets \cite{guo2016ms, cao2018vggface2, zhu2021webface260m}. FR technology has become an increasingly important application, widely used in our daily lives and even security-critical applications, such as identity checks and access control. However, the DCNN-based FR systems often involve complicated and unintuitive decision-making processes, making it difficult to interpret or further improve them. To address this problem, significant efforts have been made with the objective of enhancing the transparency and interpretability of learning-based face recognition systems. 

Early on, \cite{zhuang2010facial, kortylewski2019analyzing, terhorst2021comprehensive} reveals the bias problem of specific deep face recognition models in terms of age, gender, and ethnicity, which alerts users before launching these systems. 
More recently, saliency algorithms have become a more intuitive way of explanation for general vision models by producing heat maps highlighting regions of the input image responsible for the model's output decision. These techniques were primarily developed for explainable image classification tasks \cite{zhou2016learning, binder2016layer, ribeiro2016should, selvaraju2017grad, dabkowski2017real, chattopadhay2018grad, petsiuk2018rise}, with a few addressing other explanation problems in image retrieval \cite{hu2022x}, object detection \cite{petsiuk2021black}, etc. However, similar explanation algorithms for face recognition models are still under-explored, mainly due to the unique output formats and decision-making process. This paper aims to develop advanced saliency explanation algorithms for one of the most crucial problems in face recognition, i.e., eXplainable Face Verification (XFV), which essentially studies how a deep FR model matches a given facial image over another. 

Unlike common image classification models that often produce categorical outputs, a deep face verification system takes a pair of face images as input. It first extracts deep representation for inputs and then calculates the cosine similarity between two face embeddings. The decision is made by comparing the similarity score with a predefined threshold. Some saliency-based explanation algorithms relying on different principles have been proposed to increase the explainability of the face verification process. For example, Lin et al. \cite{lin2021xcos} plugged an external attention module to produce explainable heat maps. \cite{mery2022true, knoche2023explainable, lu2024towards} applied random perturbations to the input face images and generated saliency maps by analyzing their impact on the verification output. In \cite{williford2020explainable}, the contrastive excitation backpropagation method is used to identify the relevant salient regions on the face images. 

However, current explainable face verification approaches show some limitations. First, some methods \cite{lin2021xcos, mery2022true} only provide explanations when the FR model makes an ``Accept'' decision by presenting a similarity map while neglecting the reason for ``Reject'' decisions. 
Secondly, many popular XFV methods \cite{mery2022true, knoche2023explainable, lu2024towards} are not efficient enough and take much longer time than the verification process, hindering further practical deployment. Saliency algorithms relying on gradient backpropagation are more efficient solutions, but they often suffer from the gradient fluctuation problem and produce noisy saliency maps. To address all these issues, this paper proposes a new Feature-Guided Gradient Backpropagation (FGGB) method that provides precise and efficient explanation saliency maps for arbitrary FR systems. 
The proposed FGGB method produces both similarity and dissimilarity maps between given input images. In practice, the former leads to the explanation for ``Accept'' decisions and the latter for ``Reject'' decisions. 
Moreover, FGGB performs gradient backpropagation at feature level instead of from final scores to explore the spatial relationship between the input image and its corresponding deep feature, followed by a new saliency map generation approach that prevents the noisy gradients problem. 

\begin{figure*}[t]
\centerline{\includegraphics[width=0.8\linewidth]{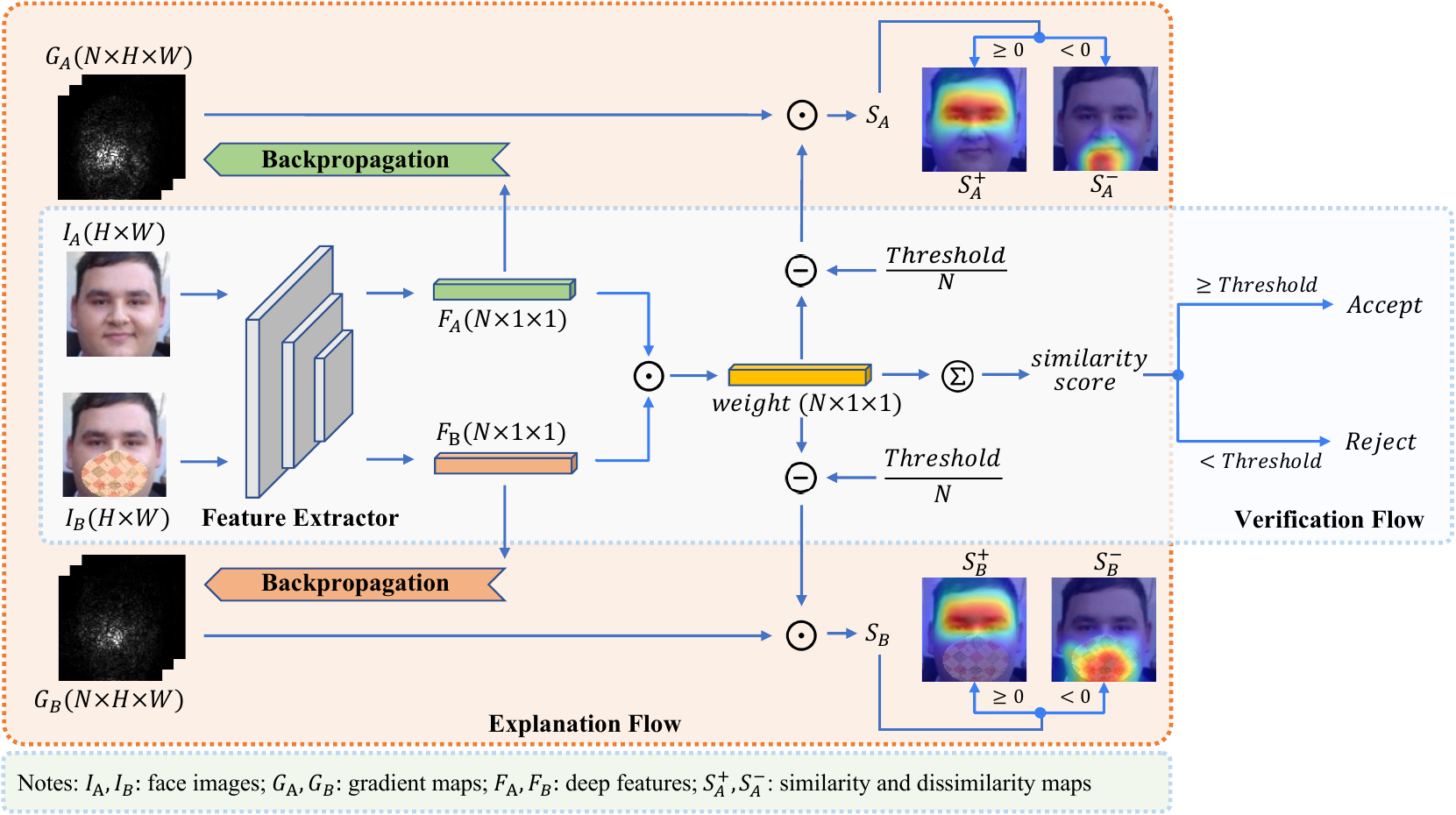}}
\caption{Workflow of the proposed Feature-Guided Gradient Backpropagation method. The similarity and dissimilarity maps are calculated respectively given an arbitrary input face pair.}
\label{fig:fggb}
\end{figure*}

\section{Related Work}
In general, most saliency-based explanation methods can be categorized into three groups based on their mechanisms. 
The first group aims to modify the internal architecture of the deep neural network to gain explainability. For example, 
CAM \cite{zhou2016learning} modified the last layer of the network. 
GAIN \cite{li2018tell} integrated learnable modules into the training process to produce attention maps. In explainable face verification, xCos \cite{lin2021xcos} added a learnable attention module to the end of the verification pipeline, and Xu et al. \cite{xu2023discriminative} trained a deep FR model together with a face reconstruction network to preserve spatial information in the face representation. However, these methods all require retraining the entire face recognition model to learn more explainable deep representation and are often not applicable to systems that have already been deployed in the real world.

The second category is so-called perturbation-based methods, which determine the salient regions by observing the effect of a perturbation on the model's output. These methods work independently from the internal status of the deep neural network and offer ``black-box'' interpretation. The idea has been popular among recent XFV methods \cite{mery2022true, knoche2023explainable, lu2024towards}. For example, Mery \cite{mery2022true} proposed to remove or aggregate different parts of images and highlight the most relevant parts for the verification process. Lu et al. \cite{lu2024towards} applied random masks to the input images and calculated both similarity and dissimilarity saliency maps through a correlation module. While these methods offer accurate and interpretable saliency maps, the perturbation-based mechanism generally lacks efficiency because they are obligated to run many iterations to guarantee a stable outcome.

Gradient backpropagation-based methods are in principle more efficient solutions. An earlier study in explainable image classification task \cite{zhang2018top} calculated the derivatives of the categorical output with respect to the input image to identify the salient pixels on the image. In XFV, Huber et al.~\cite{huber2024efficient} backpropagated the cosine similarity score between two face images to obtain saliency maps that indicate similar and dissimilar regions. However, the outcomes of this type of method tend to be noisy. SmoothGrad \cite{smilkov2017smoothgrad} for classification task sharpened saliency maps by initiatively adding noise and averaging all the resulting gradient maps. Our method takes an alternative approach to resolve this problem, by backpropagating the gradients at feature level and re-weighting the gradient maps according to the importance of each feature channel.

\section{Proposed Method}

\subsection{Problem Statement}
In principle, a face verification system makes two types of decisions, i.e., ``Accept'' and ``Reject''. This paper offers interpretations from the user's perspective and explains why the face verification system believes the given pair of facial images are matching (Accept) or non-matching (Reject). More specifically, our saliency algorithm aims to provide similarity maps for acceptance decisions and dissimilarity maps for rejection decisions. 



\subsection{Feature-Guided Gradient Backpropagation (FGGB)}
This paper proposes to leverage the gradient backpropagation method to calculate the saliency maps. 
Previous work~\cite{huber2024efficient} added a cosine similarity layer and directly backpropagated the output similarity score between two input face images throughout the verification model to get a saliency map, indicating which pixels contribute to the decision. However, a limitation of conventional gradient-based algorithms is that they identify salient pixels simply by observing raw gradient values, but the derivative of the output score may fluctuate sharply at small scales and even disappear during the backpropagation \cite{smilkov2017smoothgrad}, leading to visually noisy saliency maps. 
In general, a deep face verification system relies on the direct comparison of the distance between two deep face features, and the most discriminative feature channels often dominate the final decision. The gradient value of the most important feature channel can possibly vanish due to the fluctuation, resulting in less accurate saliency explanations.


This work conceives a simple but effective propagation scheme and saliency map generation algorithm to resolve this issue, see Fig. \ref{fig:fggb}. In specific, we perform gradient backpropagation from the deep feature level in a channel-wise manner rather than from the final score, and obtain multiple gradient maps. 
Instead of converting them to a saliency map, they are used to explore the spatial relationship between the image and its deep feature, because each gradient map will spotlight certain facial regions that correspond to an individual feature channel. Finally, the gradient maps are normalized and weight-summed by the channel-wise cosine similarity between two deep face representations, which guarantees high and stable saliency value for the most discriminative feature channels. 
Given probe and gallery images $\{I_A, I_B\}$, the FR model extracts their deep face representations, each of dimension~$N$, denoted by $\{F_A, F_B\}$. In the following, we explain the construction of similarity and dissimilarity saliency maps $S_A^+$ and $S_A^-$ using the FGGB method in detail. That of $S_B^+$ and $S_B^-$ follows similarly. The proposed method consists of two phases, (i) gradient backpropagation and (ii) saliency map generation. 


In phase (i), one first backpropagates gradient from each channel of feature~$F_A$ and construct $N$ gradient maps $G_A = \{G_A^k\colon k=1,\ldots,N\}$ as
\begin{align} \label{eq:grad}
    G_A^k = \frac{\partial F_A^k}{\partial I_A},
\end{align}
where $\partial F^k_A$ represents the derivative of the $k$-th dimension of feature $F_A$. Then, one performs normalization to each $G_A^k$ to mitigate the impact of local variations (e.g., vanishing gradient during partial derivatives) and produces
\begin{align} \label{eq:gnorm}
    \Tilde{G}_A^k = \frac{|G_A^k|}{\|G_A^k\|},
\end{align}
where $|\cdot|$ denotes the absolute value and $\|\cdot\|$ denotes the Frobenius norm of a matrix.


In phase (ii), all normalized gradient maps are accumulated to generate the saliency maps. First, a weight vector is defined to be the channel-wise cosine similarity between $\{F_A, F_B\}$:
\begin{align} \label{eq:distance}
    weight &= \frac{ F_A \odot  F_B}{\|F_A\|\|F_B\|},
\end{align}
where $\odot$ denotes the elementwise product and $\|\cdot \|$ denotes the $L_2$ norm. 

Then, features with large cosine similarity values will contribute to the similarity map and otherwise dissimilarity map. More specifically, one subtracts the decision threshold from the cosine similarity vector $weight$ and computes the saliency map $S_A$ as the weighted sum of the gradient maps,  
\begin{align} \label{eq:smap}
    S_A &= \sum_{k=1}^N \Tilde{G}_A^k \cdot \Big(weight_k - \frac{threshold}{N}\Big),
\end{align}
which is then decomposed into similarity and dissimilarity maps 
\begin{align} \label{eq:smap}
    S^+_A & = S_A[S_A\ge0],\quad S^-_A = S_A[S_A<0]. 
\end{align}

\section{Experimental Results}

    

\subsection{Implementation Details}

The proposed FGGB method is based on backpropagation and does not rely on any parameter or specific network architecture. As for deep face recognition models, this paper first conducts experiments on the popular ArcFace \cite{deng2019arcface} model with iResNet-50 \cite{hu2018squeeze} backbone. 
To show the generalization ability of FGGB across various face recognition models, its explainability performance has been tested on two additional FR models with different losses or architectures, i.e., AdaFace \cite{kim2022adaface} and MobileFaceNet \cite{chen2018mobilefacenets}. 

As comparison, five state-of-the-art explainable face verification methods, namely LIME \cite{ribeiro2016should}, MinusPlus \cite{mery2022true}, xFace \cite{knoche2023explainable}, CorrRISE \cite{lu2024towards}, and xSSAB \cite{huber2024efficient}, have been launched and tested. In detail, the third-party adaptation from \cite{mery2022true} for LIME is utilized and the official codes of MinusPlus, xFace, xSSAB, and CorrRISE have been adapted to compute in batch on GPU for acceleration purposes on multiple datasets.




\begin{figure}[t]
	\centering
	\begin{adjustbox}{width=\linewidth}
    \includegraphics[]{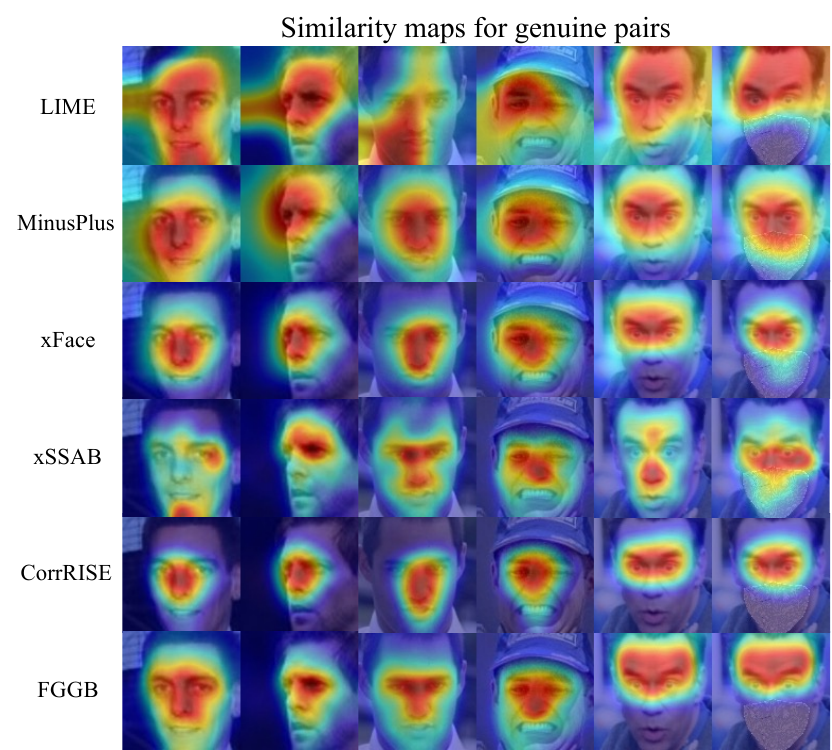}
	\end{adjustbox}
\caption{Visual comparison of similarity maps generated by FGGB and five other XFV methods. Every two columns represent a pair of genuine faces. The saliency value increases from blue to red color.}
\vspace{-4mm}
\label{fig:sim}
\end{figure}

\begin{figure}[t]
	\centering
	\begin{adjustbox}{width=\linewidth}
    \includegraphics[]{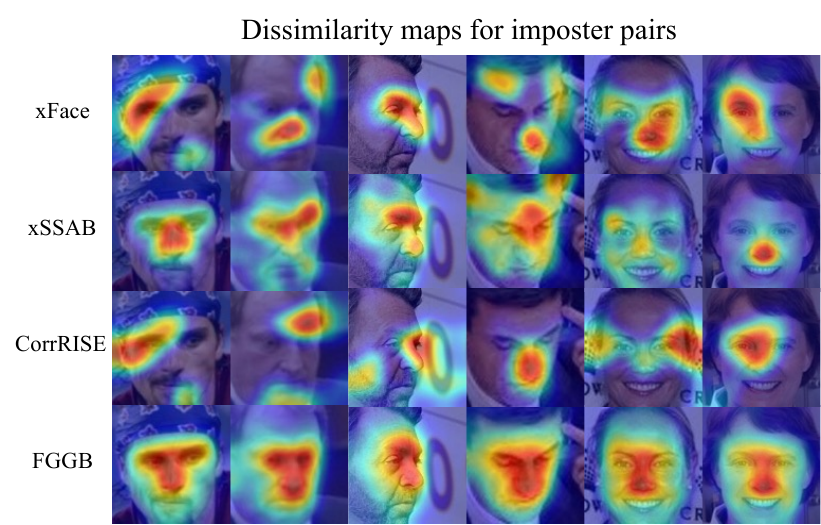}
	\end{adjustbox}
\caption{Visual comparison of dissimilarity maps generated by FGGB and other XFV methods. Every two columns represent a pair of imposter faces. The saliency value increases from blue to red color.}
\vspace{-5mm}
\label{fig:dissim}
\end{figure}

\subsection{Evaluation Methodology}
The evaluation of the proposed explanation algorithm comprises two phases. Firstly, the visualization results of produced saliency maps are presented. 
In the second phase, the evaluation metric ``Deletion\&Insertion'' proposed by \cite{lu2024towards} is employed for a quantitative comparison among state-of-the-art saliency-based XFV methods. It iteratively deletes/adds pixels from/to the inputs according to the saliency values and observes the impact on the overall verification accuracy. In general, the more precise the saliency map, the lower the ``Deletion'' score and the higher the ``Insertion'' score. However, gradient backpropagation-based methods produce more sparse salient points than perturbation-based methods and often obtain imbalanced ``Deletion'' and ``Insertion'' scores. Thus, we further improve the metric by applying a fixed-size Gaussian blur kernel to all the saliency map before evaluation, guaranteeing a fair comparison.

\begin{table}[t]
  \centering
  \caption{Quantitative evaluation of similarity maps using Deletion and Insertion metrics (\%) on LFW, CPLFW, and CALFW datasets. \textbf{Del~($\downarrow$)} refers to the Deletion metric, the smaller the better. \textbf{Ins~($\uparrow$)} refers to the Insertion metric, the larger the better. \textcolor[rgb]{ 1,  0,  0}{Red color} denotes the highest score and \textcolor[rgb]{0,0,1}{blue color} denotes the second highest score.}
    \resizebox{0.48\textwidth}{!}{
    \begin{tabular}{c|cc|cc|cc}
    \toprule
    \multirow{2}[4]{*}{Methods} & \multicolumn{2}{c|}{LFW} & \multicolumn{2}{c|}{CPLFW} & \multicolumn{2}{c}{CALFW} \\
\cmidrule{2-7}          & Del   & Ins   & Del   & Ins   & Del   & Ins \\
    \midrule
    LIME\cite{ribeiro2016should}  & 35.82 & 80.76 & 27.61 & 71.46 & 34.21 & 78.40 \\
    MinusPlus\cite{mery2022true} & 29.64 & 83.28 & 24.63 & 69.27 & 29.06 & 79.57 \\
    xFace\cite{knoche2023explainable} & 25.73 & \textcolor[rgb]{0,0,1}{86.79} & 21.82 & \textcolor[rgb]{0,0,1}{75.27} & 24.61 & \textcolor[rgb]{1,0,0}{83.24} \\
    xSSAB\cite{huber2024efficient} & 25.98 & 84.49 & 23.00 & 72.20 & 26.09 & 80.25 \\
    CorrRISE\cite{lu2024towards} & \textcolor[rgb]{ 0,0,1}{24.51} & \textcolor[rgb]{1,0,0}{86.80} & \textcolor[rgb]{1,0,0}{20.01} & \textcolor[rgb]{1,0,0}{77.07} & \textcolor[rgb]{ 0,0,1}{24.30} & \textcolor[rgb]{ 0,0,1}{83.23} \\
    FGGB & \textcolor[rgb]{1,0,0}{24.18} & 86.28 & \textcolor[rgb]{ 0,0,1}{20.25} & 74.61 & \textcolor[rgb]{1,0,0}{24.27} & 81.88 \\

    \bottomrule
    \end{tabular}%
    }
  \label{tab:sim}%
\end{table}%

\subsection{Visual Demonstration}
Given the same visualization tool, this section presents visualization results of saliency maps generated by the proposed FGGB method and other five XFV methods for face images randomly selected from CPLFW \cite{zheng2018cross}, LFW \cite{huang2008labeled}, Webface-Occ \cite{huang2021face}, and CALFW \cite{zheng2017cross} datasets, representing different verification scenarios. 

Fig. \ref{fig:sim} shows similarity maps for genuine face pairs that the FR model ``accepts''. As a result, perturbation-based methods, such as MinusPlus, xFace, and CorrRISE, tend to generate too centralized similarity maps. 
Due to the fluctuating gradient issue of propagation-based methods, xSSAB exhibits some unnatural salient regions in the first and last examples. 
In comparison, FGGB provides clear contours for decision-critical facial regions and the produced saliency maps can always accurately highlight the most similar parts between two matching faces.





On the other hand, Fig. \ref{fig:dissim} presents dissimilarity maps for imposter face pairs that the FR model ``rejects''. MinusPlus and LIME are excluded because they do not provide dissimilarity maps. It is shown that dissimilarity regions highlighted by perturbation-based methods, CorrRISE and xFace, are generally spattered at different locations and are less intuitive for interpretation. On the contrary, FGGB can produce more stable and accurate dissimilar maps for non-matching face pairs, which is also supported by later quantitative evaluation.


\begin{table}[t]
  \centering
  \caption{Quantitative evaluation of dissimilarity maps using Deletion and Insertion metrics (\%) on LFW, CPLFW, and CALFW datasets. \textbf{Del~($\downarrow$)} refers to the Deletion metric, the smaller the better. \textbf{Ins~($\uparrow$)} refers to the Insertion metric, the larger the better. \textcolor[rgb]{ 1,  0,  0}{Red color} denotes the highest score and \textcolor[rgb]{0,0,1}{blue color} denotes the second highest.}
    \resizebox{0.48\textwidth}{!}{
    \begin{tabular}{c|cc|cc|cc}
    \toprule
    \multirow{2}[4]{*}{Methods} & \multicolumn{2}{c|}{LFW} & \multicolumn{2}{c|}{CPLFW} &
    \multicolumn{2}{c}{CALFW}\\
\cmidrule{2-7}          & Del   & Ins   & Del   & Ins  & Del & Ins \\
    \midrule
    xFace\cite{knoche2023explainable} & 75.53 & 92.64 & 53.18 & 87.42 & 64.26 & \textcolor[rgb]{1,0,0}{90.91}\\
    xSSAB\cite{huber2024efficient} & \textcolor[rgb]{ 0,0,1}{49.72} & \textcolor[rgb]{1,0,0}{93.44} & \textcolor[rgb]{ 0,0,1}{33.40} & \textcolor[rgb]{ 0,0,1}{88.45} & \textcolor[rgb]{ 0,0,1}{42.03} & \textcolor[rgb]{0,0,1}{90.47} \\
    CorrRISE\cite{lu2024towards} & 81.36 & 89.81 & 57.55 & 83.98 & 69.22 & 87.34 \\
    FGGB & \textcolor[rgb]{1,0,0}{44.03} & \textcolor[rgb]{ 0,0,1}{93.35} & \textcolor[rgb]{1,0,0}{28.71} & \textcolor[rgb]{1,0,0}{88.88} & \textcolor[rgb]{1,0,0}{34.55} & 90.06 \\
    \bottomrule
    \end{tabular}%
    }
  \label{tab:dissim}%
\end{table}%

\begin{table}[t]
  \centering
  \caption{Explainablity performance of FGGB tested on different face recognition models. The verification accuracy (\%) of FR models and two explainability metrics (\%) for FGGB are reported.}
    \resizebox{\linewidth}{!}{%
    \begin{tabular}{c|c|cc}
    \toprule
    FR Models & Acc (LFW) & Deletion ($\downarrow$) & Insertion ($\uparrow$) \\
    \midrule
    ArcFace \cite{deng2019arcface} & 99.70 & 24.18 & 86.28 \\
    AdaFace \cite{kim2022adaface} & 99.27 & 21.41 & 83.87 \\
    MobileFaceNet\cite{chen2018mobilefacenets} & 98.87 & 19.32 & 77.73 \\
    \bottomrule

    \end{tabular}%
    }
  \label{tab:crossmodel}%
\end{table}%

\subsection{Quantitative Results}
Table \ref{tab:sim} quantitatively compares similarity maps obtained by FGGB and other XFV methods using the ``Deletion\&Insertion'' assessment metric. The proposed FGGB method achieves superior scores in the Deletion metric on all three datasets while getting slightly lower Insertion scores when compared to CorrRISE and xFace. As for dissimilarity maps, Table \ref{tab:dissim} shows that FGGB provides the most accurate dissimilarity maps on multiple datasets and both metrics, which is consistent with the observations in the visual demonstration. Moreover, it is notable that FGGB outperforms another propagation-based method xSSAB in most scenarios, which proves the advancement of FGGB in addressing the noisy-gradient issue during backpropagation.




In addition, FGGB is further tested on two other face recognition models with different architecture and loss functions. Table \ref{tab:crossmodel} shows that when the FR model achieves similar verification performance, the saliency maps produced by FGGB also have similar explainability performance, which validates that FGGB is model-agnostic.

\section{Conclusion}
This paper contributes to the problem of explainable face verification by conceiving a new efficient and model-agnostic saliency explanation solution FGGB. It provides similarity and dissimilarity saliency maps to interpret both the ``Accept'' and ``Reject'' decisions made by the face verification system. Experiments show that FGGB exhibits excellent performance, particularly in dissimilarity maps, when compared to the current state-of-the-art. Moreover, this paper explores a new approach to mitigate the impact of fluctuating gradients during backpropagation, which provides insights for improving future gradient propagation-based explanation methods for general learning-based vision systems.



{\small
\bibliographystyle{ieee}
\bibliography{egbib}

\begin{thebibliography}{10}\itemsep=-1pt

\bibitem{binder2016layer}
A.~Binder, G.~Montavon, S.~Lapuschkin, K.-R. M{\"u}ller, and W.~Samek.
\newblock Layer-wise relevance propagation for neural networks with local renormalization layers.
\newblock In {\em Artificial Neural Networks and Machine Learning--ICANN 2016: 25th International Conference on Artificial Neural Networks, Barcelona, Spain, September 6-9, 2016, Proceedings, Part II 25}, pages 63--71. Springer, 2016.

\bibitem{cao2018vggface2}
Q.~Cao, L.~Shen, W.~Xie, O.~M. Parkhi, and A.~Zisserman.
\newblock Vggface2: A dataset for recognising faces across pose and age.
\newblock In {\em 2018 13th IEEE international conference on automatic face \& gesture recognition (FG 2018)}, pages 67--74. IEEE, 2018.

\bibitem{chattopadhay2018grad}
A.~Chattopadhay, A.~Sarkar, P.~Howlader, and V.~N. Balasubramanian.
\newblock Grad-cam++: Generalized gradient-based visual explanations for deep convolutional networks.
\newblock In {\em 2018 IEEE winter conference on applications of computer vision (WACV)}, pages 839--847. IEEE, 2018.

\bibitem{chen2018mobilefacenets}
S.~Chen, Y.~Liu, X.~Gao, and Z.~Han.
\newblock Mobilefacenets: Efficient cnns for accurate real-time face verification on mobile devices.
\newblock In {\em Biometric Recognition: 13th Chinese Conference, CCBR 2018, Urumqi, China, August 11-12, 2018, Proceedings 13}, pages 428--438. Springer, 2018.

\bibitem{dabkowski2017real}
P.~Dabkowski and Y.~Gal.
\newblock Real time image saliency for black box classifiers.
\newblock {\em Advances in neural information processing systems}, 30, 2017.

\bibitem{deng2019arcface}
J.~Deng, J.~Guo, N.~Xue, and S.~Zafeiriou.
\newblock Arcface: Additive angular margin loss for deep face recognition.
\newblock In {\em Proceedings of the IEEE/CVF conference on computer vision and pattern recognition}, pages 4690--4699, 2019.

\bibitem{guo2016ms}
Y.~Guo, L.~Zhang, Y.~Hu, X.~He, and J.~Gao.
\newblock Ms-celeb-1m: A dataset and benchmark for large-scale face recognition.
\newblock In {\em Computer Vision--ECCV 2016: 14th European Conference, Amsterdam, The Netherlands, October 11-14, 2016, Proceedings, Part III 14}, pages 87--102. Springer, 2016.

\bibitem{he2016deep}
K.~He, X.~Zhang, S.~Ren, and J.~Sun.
\newblock Deep residual learning for image recognition.
\newblock In {\em Proceedings of the IEEE conference on computer vision and pattern recognition}, pages 770--778, 2016.

\bibitem{hu2022x}
B.~Hu, B.~Vasu, and A.~Hoogs.
\newblock X-mir: Explainable medical image retrieval.
\newblock In {\em Proceedings of the IEEE/CVF Winter Conference on Applications of Computer Vision}, pages 440--450, 2022.

\bibitem{hu2018squeeze}
J.~Hu, L.~Shen, and G.~Sun.
\newblock Squeeze-and-excitation networks.
\newblock In {\em Proceedings of the IEEE conference on computer vision and pattern recognition}, pages 7132--7141, 2018.

\bibitem{huang2021face}
B.~Huang, Z.~Wang, G.~Wang, K.~Jiang, K.~Zeng, Z.~Han, X.~Tian, and Y.~Yang.
\newblock When face recognition meets occlusion: A new benchmark.
\newblock In {\em ICASSP 2021-2021 IEEE International Conference on Acoustics, Speech and Signal Processing (ICASSP)}, pages 4240--4244. IEEE, 2021.

\bibitem{huang2008labeled}
G.~B. Huang, M.~Mattar, T.~Berg, and E.~Learned-Miller.
\newblock Labeled faces in the wild: A database forstudying face recognition in unconstrained environments.
\newblock In {\em Workshop on faces in'Real-Life'Images: detection, alignment, and recognition}, 2008.

\bibitem{huber2024efficient}
M.~Huber, A.~T. Luu, P.~Terh{\"o}rst, and N.~Damer.
\newblock Efficient explainable face verification based on similarity score argument backpropagation.
\newblock In {\em Proceedings of the IEEE/CVF Winter Conference on Applications of Computer Vision}, pages 4736--4745, 2024.

\bibitem{kim2022adaface}
M.~Kim, A.~K. Jain, and X.~Liu.
\newblock Adaface: Quality adaptive margin for face recognition.
\newblock In {\em Proceedings of the IEEE/CVF Conference on Computer Vision and Pattern Recognition}, pages 18750--18759, 2022.

\bibitem{knoche2023explainable}
M.~Knoche, T.~Teepe, S.~H{\"o}rmann, and G.~Rigoll.
\newblock Explainable model-agnostic similarity and confidence in face verification.
\newblock In {\em Proceedings of the IEEE/CVF Winter Conference on Applications of Computer Vision}, pages 711--718, 2023.

\bibitem{kortylewski2019analyzing}
A.~Kortylewski, B.~Egger, A.~Schneider, T.~Gerig, A.~Morel-Forster, and T.~Vetter.
\newblock Analyzing and reducing the damage of dataset bias to face recognition with synthetic data.
\newblock In {\em Proceedings of the IEEE/CVF Conference on Computer Vision and Pattern Recognition Workshops}, pages 0--0, 2019.

\bibitem{li2018tell}
K.~Li, Z.~Wu, K.-C. Peng, J.~Ernst, and Y.~Fu.
\newblock Tell me where to look: Guided attention inference network.
\newblock In {\em Proceedings of the IEEE conference on computer vision and pattern recognition}, pages 9215--9223, 2018.

\bibitem{lin2021xcos}
Y.-S. Lin, Z.-Y. Liu, Y.-A. Chen, Y.-S. Wang, Y.-L. Chang, and W.~H. Hsu.
\newblock xcos: An explainable cosine metric for face verification task.
\newblock {\em ACM Transactions on Multimedia Computing, Communications, and Applications (TOMM)}, 17(3s):1--16, 2021.

\bibitem{lu2024towards}
Y.~Lu, Z.~Xu, and T.~Ebrahimi.
\newblock Towards visual saliency explanations of face verification.
\newblock In {\em Proceedings of the IEEE/CVF Winter Conference on Applications of Computer Vision}, pages 4726--4735, 2024.

\bibitem{mery2022true}
D.~Mery.
\newblock True black-box explanation in facial analysis.
\newblock In {\em Proceedings of the IEEE/CVF Conference on Computer Vision and Pattern Recognition}, pages 1596--1605, 2022.

\bibitem{petsiuk2018rise}
V.~Petsiuk, A.~Das, and K.~Saenko.
\newblock Rise: Randomized input sampling for explanation of black-box models.
\newblock {\em arXiv preprint arXiv:1806.07421}, 2018.

\bibitem{petsiuk2021black}
V.~Petsiuk, R.~Jain, V.~Manjunatha, V.~I. Morariu, A.~Mehra, V.~Ordonez, and K.~Saenko.
\newblock Black-box explanation of object detectors via saliency maps.
\newblock In {\em Proceedings of the IEEE/CVF Conference on Computer Vision and Pattern Recognition}, pages 11443--11452, 2021.

\bibitem{ribeiro2016should}
M.~T. Ribeiro, S.~Singh, and C.~Guestrin.
\newblock " why should i trust you?" explaining the predictions of any classifier.
\newblock In {\em Proceedings of the 22nd ACM SIGKDD international conference on knowledge discovery and data mining}, pages 1135--1144, 2016.

\bibitem{selvaraju2017grad}
R.~R. Selvaraju, M.~Cogswell, A.~Das, R.~Vedantam, D.~Parikh, and D.~Batra.
\newblock Grad-cam: Visual explanations from deep networks via gradient-based localization.
\newblock In {\em Proceedings of the IEEE international conference on computer vision}, pages 618--626, 2017.

\bibitem{simonyan2014very}
K.~Simonyan and A.~Zisserman.
\newblock Very deep convolutional networks for large-scale image recognition.
\newblock {\em arXiv preprint arXiv:1409.1556}, 2014.

\bibitem{smilkov2017smoothgrad}
D.~Smilkov, N.~Thorat, B.~Kim, F.~Vi{\'e}gas, and M.~Wattenberg.
\newblock Smoothgrad: removing noise by adding noise.
\newblock {\em arXiv preprint arXiv:1706.03825}, 2017.

\bibitem{tan2019efficientnet}
M.~Tan and Q.~Le.
\newblock Efficientnet: Rethinking model scaling for convolutional neural networks.
\newblock In {\em International conference on machine learning}, pages 6105--6114. PMLR, 2019.

\bibitem{terhorst2021comprehensive}
P.~Terh{\"o}rst, J.~N. Kolf, M.~Huber, F.~Kirchbuchner, N.~Damer, A.~M. Moreno, J.~Fierrez, and A.~Kuijper.
\newblock A comprehensive study on face recognition biases beyond demographics.
\newblock {\em IEEE Transactions on Technology and Society}, 3(1):16--30, 2021.

\bibitem{williford2020explainable}
J.~R. Williford, B.~B. May, and J.~Byrne.
\newblock Explainable face recognition.
\newblock In {\em Computer Vision--ECCV 2020: 16th European Conference, Glasgow, UK, August 23--28, 2020, Proceedings, Part XI}, pages 248--263. Springer, 2020.

\bibitem{xu2023discriminative}
Z.~Xu, Y.~Lu, and T.~Ebrahimi.
\newblock Discriminative deep feature visualization for explainable face recognition.
\newblock {\em arXiv preprint arXiv:2306.00402}, 2023.

\bibitem{zhang2018top}
J.~Zhang, S.~A. Bargal, Z.~Lin, J.~Brandt, X.~Shen, and S.~Sclaroff.
\newblock Top-down neural attention by excitation backprop.
\newblock {\em International Journal of Computer Vision}, 126(10):1084--1102, 2018.

\bibitem{zheng2018cross}
T.~Zheng and W.~Deng.
\newblock Cross-pose lfw: A database for studying cross-pose face recognition in unconstrained environments.
\newblock {\em Beijing University of Posts and Telecommunications, Tech. Rep}, 5:7, 2018.

\bibitem{zheng2017cross}
T.~Zheng, W.~Deng, and J.~Hu.
\newblock Cross-age lfw: A database for studying cross-age face recognition in unconstrained environments.
\newblock {\em arXiv preprint arXiv:1708.08197}, 2017.

\bibitem{zhou2016learning}
B.~Zhou, A.~Khosla, A.~Lapedriza, A.~Oliva, and A.~Torralba.
\newblock Learning deep features for discriminative localization.
\newblock In {\em Proceedings of the IEEE conference on computer vision and pattern recognition}, pages 2921--2929, 2016.

\bibitem{zhu2021webface260m}
Z.~Zhu, G.~Huang, J.~Deng, Y.~Ye, J.~Huang, X.~Chen, J.~Zhu, T.~Yang, J.~Lu, D.~Du, et~al.
\newblock Webface260m: A benchmark unveiling the power of million-scale deep face recognition.
\newblock In {\em Proceedings of the IEEE/CVF Conference on Computer Vision and Pattern Recognition}, pages 10492--10502, 2021.

\bibitem{zhuang2010facial}
Z.~Zhuang, D.~Landsittel, S.~Benson, R.~Roberge, and R.~Shaffer.
\newblock Facial anthropometric differences among gender, ethnicity, and age groups.
\newblock {\em Annals of occupational hygiene}, 54(4):391--402, 2010.

\end{thebibliography}
}

\end{document}